\documentstyle[11pt,aclap]{article}

  \catcode`\  = \active                \def {~}       
  \catcode`\¢ = \active                \def¢{\u{ }}            
  \catcode`\£ = \active                \def£{{\L}}                
  \catcode`\¥ = \active                \def¥{\v{L}}               
  \catcode`\¦ = \active                \def¦{\'S}                 
  \catcode`\§ = \active                \def§{\S}                  
  \catcode`\¨ = \active                \def¨{\"{ }}               
  \catcode`\© = \active                \def©{\v{S}}               
  \catcode`\ª = \active                \defª{\c{S}}               
  \catcode`\« = \active                \def«{\v{T}}               
  \catcode`\¬ = \active                \def¬{\'Z}                 
  \catcode`\® = \active                \def®{\v{Z}}               
  \catcode`\¯ = \active                \def¯{\.{Z}}               
  \catcode`\° = \active                \def°{{\accent'27 }}       
  \catcode`\³ = \active                \def³{{\l}}                
  \catcode`\´ = \active                \def´{\'{ }}              
  \catcode`\µ = \active                \defµ{\v{l}}               
  \catcode`\¶ = \active                \def¶{\'s}                 
  \catcode`\· = \active                \def·{\v{ }}             
  \catcode`\¸ = \active                \def¸{\c{ }}             
  \catcode`\¹ = \active                \def¹{\v{s}}               
  \catcode`\º = \active                \defº{\c{s}}               
  \catcode`\» = \active                \def»{\v{t}}               
  \catcode`\¼ = \active                \def¼{\'z}                 
  \catcode`\½ = \active                \def½{\H{ }}             
  \catcode`\¾ = \active                \def¾{\v{z}}               
  \catcode`\¿ = \active                \def¿{\.{z}}               
  \catcode`\À = \active                \defÀ{\'R}                 
  \catcode`\Á = \active                \defÁ{\'A}                 
  \catcode`\Â = \active                \defÂ{\^A}                 
  \catcode`\Ã = \active                \defÃ{\u{A}}               
  \catcode`\Ä = \active                \defÄ{\"A}                 
  \catcode`\Å = \active                \defÅ{\'L}                 
  \catcode`\Æ = \active                \defÆ{\'C}                 
  \catcode`\Ç = \active                \defÇ{\c{C}}               
  \catcode`\È = \active                \defÈ{\v{C}}               
  \catcode`\É = \active                \defÉ{\'E}                 
  \catcode`\Ë = \active                \defË{\"E}                 
  \catcode`\Ì = \active                \defÌ{\v{E}}               
  \catcode`\Í = \active                \defÍ{\'I}                 
  \catcode`\Î = \active                \defÎ{\^I}                 
  \catcode`\Ï = \active                \defÏ{\v{D}}               
  \catcode`\Ñ = \active                \defÑ{\'N}                 
  \catcode`\Ò = \active                \defÒ{\v{N}}               
  \catcode`\Ó = \active                \defÓ{\'O}                 
  \catcode`\Ô = \active                \defÔ{\^O}                 
  \catcode`\Õ = \active                \defÕ{\H{O}}               
  \catcode`\Ö = \active                \defÖ{\"O}                 
  \catcode`\× = \active                \def×{$\times$}              
  \catcode`\Ø = \active                \defØ{\v{R}}               
  \catcode`\Ù = \active                \defÙ{{\accent'27U}}       
  \catcode`\Ú = \active                \defÚ{\'U}                 
  \catcode`\Û = \active                \defÛ{\H{U}}               
  \catcode`\Ü = \active                \defÜ{\"U}                 
  \catcode`\Ý = \active                \defÝ{\'Y}                 
  \catcode`\Þ = \active                \defÞ{\c{T}}               
  \catcode`\ß = \active                \defß{{\ss}}               
  \catcode`\à = \active                \defà{\'r}                 
  \catcode`\á = \active                \defá{\'a}                 
  \catcode`\â = \active                \defâ{\^a}                 
  \catcode`\ã = \active                \defã{\u{a}}               
  \catcode`\ä = \active                \defä{\"a}                 
  \catcode`\å = \active                \defå{\'l}                 
  \catcode`\æ = \active                \defæ{\'c}                 
  \catcode`\ç = \active                \defç{\c{c}}               
  \catcode`\è = \active                \defè{\v{c}}               
  \catcode`\é = \active                \defé{\'e}                 
  \catcode`\ë = \active                \defë{\"e}                 
  \catcode`\ì = \active                \defì{\v{e}}               
  \catcode`\í = \active                \defí{\'{\i}}              
  \catcode`\î = \active                \defî{\^i}                 
  \catcode`\ï = \active                \defï{\v{d}}               
  \catcode`\ñ = \active                \defñ{\'n}                 
  \catcode`\ò = \active                \defò{\v{n}}               
  \catcode`\ó = \active                \defó{\'o}                 
  \catcode`\ô = \active                \defô{\^o}                 
  \catcode`\õ = \active                \defõ{\H{o}}               
  \catcode`\ö = \active                \defö{\"o}                 
  \catcode`\÷ = \active                \def÷{$\div$}                
  \catcode`\ø = \active                \defø{\v{r}}               
  \catcode`\ù = \active                \defù{{\accent'27u}}       
  \catcode`\ú = \active                \defú{\'u}                 
  \catcode`\û = \active                \defû{\H{u}}               
  \catcode`\ü = \active                \defü{\"u}                 
  \catcode`\ý = \active                \defý{\'y}                 
  \catcode`\þ = \active                \defþ{\c{t}}               
  \catcode`\ÿ = \active                \defÿ{\.{ }} 

\title{A Czech Morphological Lexicon} \author{Hana Skoumalová \\
Institute of Theoretical and Computational Linguistics \\ Charles
University \\ Celetná 13, Praha 1 \\ Czech Republic \\ {\em
hana.skoumalova@ff.cuni.cz}}

\begin{document}

\maketitle

\bibliographystyle{acl}

\begin{abstract}
In this paper, a treatment of Czech phonological rules in two-level
morphology approach is described. First the possible phonological
alternations in Czech are listed and then their treatment in a
practical application of a Czech morphological lexicon.
\end{abstract}

\section{Motivation}

In this paper I want to describe the way in which I treated the
phonological changes that occur in Czech conjugation, declension and
derivation. My work concerned the written language, but as spelling of
Czech is based on phonological principles, most statements will be
true about phonology, too.

My task was to encode an existing Czech morphological
dictionary~\cite{Haj:mor} as a finite state transducer. The existing
lexicon was originally designed for simple C programs that only attach
``endings'' to the ``stems''. The quotation marks in the previous
sentence mean that the terms are not used in the linguistic meaning
but rather technically: {\em Stem} means any part of a word that is
not changed in declension/conjugation. {\em Ending} means the real
ending and possibly also another part of the word that is changed.
When I started the work on converting this lexicon to a two-level
morphology system, the first idea was that it should be linguistically
more elegant and accurate. This required me to redesign the set of
patterns and their corresponding endings. From the original number of
219 paradigms I got 159 that use 116 sets of endings. Under the term
paradigm I mean the set of endings that belong to one lemma (e.g. noun
endings for all seven cases in both numbers) and possible derivations
with their corresponding endings (e.g. possessive adjectives derived
from nouns in all possible forms). That is why the number of paradigms
is higher then the number of endings.

In this approach, it is necessary to deal with the phonological
changes that occur at boundaries between the stem and the
suffix/ending or between the suffix and the ending. There are also
changes inside the stem (e.g. {\em pøítel} `friend' $\times$
{\em pøátelé} `friends', or {\em hnát} `to chase'
$\times$ {\em ¾enu} `I chase'), but I will not deal with them, as
they are rather rare and irregular. They are treated in the lexicon as
exceptions. I also will not deal with all the changes that may occur
in a verb stem---this would require reconstructing the forms of the
verbs back in the 14th century, which is outside the scope of my
work. Instead, I work with several stems of these irregular verbs. For
example the verb {\em hnát} (`to chase') has three different
stems, {\em hná-} for infinitive, {\em ¾en-} for the present
tense, imperative and present participles, and {\em hna-} for the past
participles. The verb {\em vést} (`to lead') has two stems, {\em
vés-} for the infinitive and {\em ved-} for all finite forms and
participles. The verb {\em tít} (`to cut') has the stem {\em tn-}
in the present tense, and the stem {\em »a-} in the past tense; the
participles can be formed both from the present and the past stem.
For practical reasons we work either with one verb stem (for regular
verbs) or with six stems (for irregular verbs). These six stems are
stems for infinitive, present indicative, imperative, past participle,
transgressive and passive participle. In fact, there is no verb in
Czech with six different stems, but this division is made because of
various combinations of endings with the stems.

\section{Types of phonological alternations in Czech}

We will deal with three types of phonological alternations:
palatalization, assimilation and epenthesis. Palatalization occurs
mainly in declension and partly also in conjugation. Assimilation
occurs mainly in conjugation. Epenthesis occurs both in declension
and in conjugation.

\subsection{Epenthesis}

An epenthetic {\em e} occurs in a group of consonants before a
\O-ending. The final group of consonants can consist of a suffix
(e.g. {\em -k} or {\em -b}) and a part of the stem; in this case the
epenthesis is obligatory (e.g. {\em kousek} $\times$ {\em kousku}
`piece', {\em malba} $\times$ {\em maleb} `painting'). In cases when
the group is morphologically unseparable, the application of
epenthesis depends on whether the group of consonants is phonetically
admissable at word end. In loan words, the epenthetic {\em e} may occur
if the final group of consonants reminds a Czech suffix (e.g. {\em
korek} $\times$ {\em korku} `cork', but {\em alba} $\times$ {\em
alb} `alb'). In declension, two situations can occur:

\begin{itemize}
\item The base form contains an epenthetic {\em e}; the rule has to
remove it, if the form has a non-{\O} ending, e.g. {\em chlapec} `boy',
{\em chlapci} dative/locative sg or nominative pl.
\item The base form has a non-{\O} ending; the rule has to insert an
epenthetic {\em e}, if the ending is \O, e.g. {\em chodba}
`corridor', {\em chodeb} genitive pl.
\end{itemize}

In conjugation, an epenthetic {\em e} occurs in the past participle,
masculine sg of the verb {\em jít} `to go' (and its prefixed
derivations): {\em ¹el} `he-gone', {\em ¹la} `she-gone', {\em ¹lo}
`it-gone'. The rule has to insert an epenthetic {\em e} if the form
has a \O-ending.

\subsection{Palatalization and assimilation}

Palatalization or assimilation at the morpheme boundaries occurs when
an ending/suffix starts with a soft vowel. The alternations are
different for different types of consonants. The types of consonants
and vowels are as follows:

\begin{itemize}
\item hard consonants---{\em d, (g,) h, ch, k, n, r, t}
\item soft consonants---{\em c, è, ï, j, ò, ø, ¹, », ¾}
\item neutral consonants---{\em b, l, m, p, s, v, z}
\item hard vowels---{\em a, á, e, é, o, ó, u, ù, y, ý} and the
diphthong {\em ou}
\item soft vowels---{\em ì, i, í}
\end{itemize}

The vowel {\em ú} cannot occur in the ending/suffix so it will not be
interesting for us. I also will not discuss what happens with
`foreign' consonants {\em f, q, w} and {\em x}---they would be treated
as {\em v, k, v} and {\em s}, respectively. The only borrowing from
foreign languages that I included to the above lists is {\em g}: This
sound existed in Old Slavonic but in Czech it changed into {\em
h}. However, when later new words with {\em g} were adopted from other
languages, this sound behaved phonologically as {\em h} (e.g. {\em
hloh, hlozích}---from Common Slavonic {\em glog} `hawthorn', and {\em
katalog, katalozích} `catalog').

The phonological alternations are reflected in writing, with one
exception---if the consonants {\em d, n} and {\em t} are followed by a
soft vowel, they are palatalized, but the spelling is not changed:

\begin{tabbing}
spelling: \= {\em dì, di} mezera \= phonology: \= \kill spelling:
\> {\em dì, di} \> phonology: \> /{\em ïe}/, /{\em ïi}/ \\
\> {\em nì, ni} \> \> /{\em òe}/, /{\em òi}/ \\ \> {\em
tì, ti} \> \> /{\em »e}/, /{\em »i}/
\end{tabbing}

In other cases the spelling reflects the phonology. In the further
text I will use \{~\} for the morpho-phonological level, /~/ for the
phonological level and no brackets for the orthographical level. In
the cases where the orthography and phonology are the same I will only
use the orthographical level. Let us look at the possible types of
alternation of consonants:

\begin{itemize}
\item Soft consonant and {\em ì} --- The soft consonant is not
changed, the soft {\em ì} is changed to {\em e}. \\ \{{\em èíèì}\}
$\rightarrow$ {\em èíèe} `pussycat' dative sg

\item Soft or neutral consonant and {\em i/í} --- No alternations
occur. \\ \{{\em èíèi}\} $\rightarrow$ {\em èíèi} `pussycat' genitive
sg

\item Hard consonant and a soft vowel --- The alternations differ
depending on when and how the soft vowel originated. \\ \\
Assimilation:
\begin{itemize}
\item \{{\em kj}\} $\rightarrow$ {\em è} \\ {\em tlak} `pressure'
$\rightarrow$ {\em tlaèen} `pressed'
\item \{{\em hj}\} $\rightarrow$ {\em ¾} \\ {\em mnoho} `much, many'
$\rightarrow$ {\em mno¾ení} `multiplying'
\item \{{\em gj}\} $\rightarrow$ {\em ¾} \\ It is not easy to find an
example of this sort of alternation, as {\em g} only occurs in loan
words that do not use the old types of derivation. In colloquial
speech it would be perhaps possible to create the following form: \\
{\em pedagog} `teacher' $\rightarrow$ {\em pedago¾ení} `working as a
teacher'
\item \{{\em dj}\} $\rightarrow$ {\em z} \\ {\em sladit} `to sweeten'
$\rightarrow$ {\em slazení} `sweetening' \\ This sort of alternation
is not productive any more---in newer words palatalization applies: \\
{\em sladit} `to tune up' $\rightarrow$ {\em sladìní} `tuning up' \\
In some cases both variants are possible, or the different variants
exist in different dialects---the east (Moravian) dialects tend to
keep this phonological alternation, while the west (Bohemian) dialects
often abandoned it.
\item \{{\em tje}\} $\rightarrow$ {\em ce} \\ {\em platit} `to pay'
$\rightarrow$ {\em placení} `paying' \\ This alternation is also not
productive any more. The newest word that I found which shows this
sort of phonological alternation is the word {\em fotit} `to take a
photo' $\rightarrow$ {\em focení} `taking a photo'.
\end{itemize}
Palatalization: \\ During the historical development of the language
several sorts of palatalization occured---the first and second
Slavonic palatalization and further Czech palatalizations.
\begin{itemize}
\item \{{\em kì/ki}\} $\rightarrow$ {\em èe/èi} (1st palat.) \\ {\em
matka} `mother' $\rightarrow$ {\em matèin} possesive adjective
\item \{{\em kì/ki}\} $\rightarrow$ {\em ce/ci} (2nd palat.) \\ {\em
matka} $\rightarrow$ {\em matce} dative/locative sg
\item \{{\em hì/hi}\} $\rightarrow$ {\em ¾e/¾i} (1st palat.) \\ {\em
Bùh} `God' $\rightarrow$ {\em Bo¾e} vocative sg
\item \{{\em hì/hi}\} $\rightarrow$ {\em ze/zi} (2nd palat.) \\ {\em
Bùh} $\rightarrow$ {\em Bozi} nominative/vocative pl
\item \{{\em gì/gi}\} $\rightarrow$ {\em ¾e/¾i} (1st palat.) \\ {\em
Jaga} a witch from Russian tales $\rightarrow$ {\em Ja¾in} possesive
adjective
\item \{{\em gì/gi}\} $\rightarrow$ {\em ze/zi} (2nd palat.) \\ {\em
Jaga} $\rightarrow$ {\em Jaze} dative/locative sg
\item \{{\em dì}\} $\rightarrow$ /{\em ïe}/ $\rightarrow$ {\em dì} \\
{\em rada} `council' $\rightarrow$ {\em radì} dative/locative sg
\item \{{\em tì}\} $\rightarrow$ /{\em »e}/ $\rightarrow$ {\em tì} \\
{\em teta} `aunt' $\rightarrow$ {\em tetì} dative/locative sg
\end{itemize}
Both palatalization and assimilation yields the same result:
\begin{itemize}
\item \{{\em ch}\} $\rightarrow$ {\em ¹} \\ {\em moucha} `fly'
$\rightarrow$ {\em mou¹e} dative/locative sg, {\em mu¹í} derived
adjective
\item \{{\em n}\} $\rightarrow$ /{\em ò}/ $\rightarrow$ {\em n} \\
{\em hon} `chase' $\rightarrow$ {\em honit} `to chase', {\em honìný}
`chased'
\item \{{\em r}\} $\rightarrow$ {\em ø} \\ {\em var} `boil'
$\rightarrow$ {\em vaøit} `to cook', {\em vaøení} `cooking'
\end{itemize}
\item Neutral consonant and {\em ì} --- The alternations differ
depending on when and how {\em ì} originated. \\ \\ Assimilation:
\begin{itemize}
\item \{{\em bje}\} $\rightarrow$ {\em be} \\ {\em zlobit} `to
irritate' $\rightarrow$ \{{\em zlobjení}\} $\rightarrow$ {\em zlobení}
`irritating'
\item \{{\em mje}\} $\rightarrow$ {\em me} \\ {\em zlomit} `to break'
$\rightarrow$ \{{\em zlomjený}\} $\rightarrow$ {\em zlomený} `broken'
\item \{{\em pje}\} $\rightarrow$ {\em pe} \\ {\em kropit} `to
sprinkle' $\rightarrow$ \{{\em kropjení}\} $\rightarrow$ {\em kropení}
`sprinkling'
\item \{{\em vje}\} $\rightarrow$ {\em ve} \\ {\em lovit} `to hunt'
$\rightarrow$ \{{\em lovjení}\} $\rightarrow$ {\em lovení} `hunting'
\item \{{\em sje}\} $\rightarrow$ {\em ¹e} \\ {\em prosit} `to ask'
$\rightarrow$ \{{\em prosjení}\} $\rightarrow$ {\em pro¹ení} `asking'
\\ This type of assimilation is not productive any more. In newer
derivations \{{\em sje}\} $\rightarrow$ {\em se} (e.g. {\em kosit} `to
mow' $\rightarrow$ {\em kosení} `mowing').
\item \{{\em zje}\} $\rightarrow$ {\em ¾e} \\ {\em kazit} `to spoil'
$\rightarrow$ \{{\em kazjení}\} $\rightarrow$ {\em ka¾ení} `spoiling'
\\ This type of assimilation is also not productive any more. In newer
derivations \{{\em zje}\} $\rightarrow$ {\em ze} (e.g. {\em øetìzit}
`to concatenate' $\rightarrow$ {\em øetìzení} `concatenating').
\end{itemize}
Palatalization: \\ With {\em b, m, p} and {\em v} no alternation
occurs (\{{\em vrbì}\} `willow' dative/locative sg $\rightarrow$ {\em
vrbì}).
\begin{itemize}
\item \{{\em sì}\} $\rightarrow$ {\em se} \\ {\em vosa} `wasp'
$\rightarrow$ \{{\em vosì}\} $\rightarrow$ {\em vose} dative/locative
sg
\item \{{\em zì}\} $\rightarrow$ {\em ze} \\ {\em koza} `goat'
$\rightarrow$ \{{\em kozì}\} $\rightarrow$ {\em koze} dative/locative
sg
\end{itemize}
Both palatalization and assimilation yields the same result:
\begin{itemize}
\item \{{\em lje}\} $\rightarrow$ {\em le} \\ {\em ¹kolit} `to school'
$\rightarrow$ \{{\em ¹koljení}\} $\rightarrow$ {\em ¹kolení} `schooling'
\item \{{\em lì}\} $\rightarrow$ {\em le} \\ {\em ¹kola} `school'
$\rightarrow$ \{{\em ¹kolì}\} $\rightarrow$ {\em ¹kole} dative/locative
sg
\end{itemize}

\item Group of hard consonants and a soft vowel. Here again either
palatalization or assimilation occurs. \\ \\ Assimilation:
\begin{itemize}
\item \{{\em stj}\} $\rightarrow$ /{\em ¹»}/ \\
{\em èistit} `to clean' $\rightarrow$ {\em èi¹tìní} `cleaning'
\item \{{\em slj}\} $\rightarrow$ {\em ¹l} \\
{\em myslit} `to think' $\rightarrow$ {\em my¹lení} `thinking'
\end{itemize}
Palatalization:
\begin{itemize}
\item \{{\em sk}\} $\rightarrow$ /{\em ¹»}/ \\
{\em kamarádský} `friendly' $\rightarrow$ {\em kamarád¹tí} masculine
animate, nominative pl, {\em kamarád¹tìj¹í} `more friendly'
\item \{{\em ck}\} $\rightarrow$ /{\em è»}/ \\
{\em èacký} `brave' $\rightarrow$ {\em èaètí} masculine
animate, nominative pl, {\em èaètìj¹í} `braver'
\item \{{\em èk}\} $\rightarrow$ /{\em è»}/ \\ {\em ¾lu»ouèký}
`yellowish' $\rightarrow$ {\em ¾lu»ouètìj¹í} `more yellowish', but
{\em ¾lu»ouècí} masculine animate, nominative pl
\end{itemize}
\end{itemize}

The alternations affect also the vowel {\em ì}. When it causes
palatalization or assimilation of the previous consonant, it looses
its `softness', i.e. {\em ì} $\rightarrow$ {\em e}: \\ \{{\em matkì}\}
$\rightarrow$ {\em matce} \\ \{{\em sestrì}\} $\rightarrow$ {\em sestøe}
\\ \{{\em ¹kolì}\} $\rightarrow$ {\em ¹kole}

\section{Phenomena treated by two-level rules in the Czech lexicon}

As the Czech lexicon should serve practical applications I did not try
to solve all the problems that occur in Czech phonology. I
concentrated on dealing with the alternations that occur in declension
and regular conjugation, and the most productive derivations. The rest
of alternations occurring in conjugation are treated by inserting
several verb stems in the lexicon. The list of alternations and other
changes covered by the rules:

\begin{itemize}
\item epenthesis
\item palatalization in declension
\item palatalization in conjugation
\item palatalization in derivation of feminine nouns from masculines
\item palatalization in derivation of possessive adjectives
\item palatalization in derivation of adverbs
\item palatalization in derivation of comparatives of adjectives and
adverbs
\item palatalization or assimilation in derivation of passive
participles
\item shortening of the vowel in suffixes {\em -ík} (in derivation of
feminine noun from masculine) and {\em -ùv} (in declension of
possesive adjectives)
\end{itemize}

For the Czech lexicon I used the software tools for two-level
morphology developed at Xerox~\cite{X:twolc,X:lexc}. The lexical forms
are created by attaching the proper ending/suffix to the base form in
a separate program. To help the two-level rules to find where they
should operate, I also marked morpheme boundaries by special
markers. These markers have two further functions:

\begin{itemize}
\item They bear the information about the length of ending (or suffix
and ending) of the base form, i.e. how many characters should be
removed before attaching the ending.
\item They bear the information about the kind of alternation.
\end{itemize}

Beside the markers for morpheme boundaries I also use markers for an
epenthetic {\em e}. As I said before, {\em e} is inserted before
the last consonat of a final consonant group, if the last consonant is
a suffix, or if the consonant group is not phonetically admissable.
However, as I do not generally deal with derivation nor with the
phonetics, I am not able to recognize what is a suffix and what is
phonetically admissable.  That is why I need these special markers.

Another auxiliary marker is used for marking the suffix {\em -ík},
that needs a special treatment in derivation of feminine nouns and
their possesive adjectives.  The long vowel {\em í} must be shortened
in the derivation, and the final {\em k} must be palatalized even if
the \O-ending follows. I need a special marker, as {\em -ík-} allows
two realizations for both the sounds in same contexts: \\ Two
realizations of {\em í} \\ {\em úøedník} `clerk' $\rightarrow$ {\em
úøednice} `she-clerk', but {\em rybník} `pond' $\rightarrow$ {\em
rybníce} locative sg \\ Two realizations of {\em k} \\ {\em úøedník}
$\times$ {\em úøednic} (genitive pl of the derived feminine)

In the previous section, I described all possible alternations
concerning single consonants. When I work with the paradigms or with
the derivations, it is necessary to specify the kind of the
alternation for all consonants that can occur at the
boundary. For this purpose I introduced four types of
markers:

\begin{description}
\item[\^{}1P] --- 1st palatalization for {\em g, h} and {\em k}, or
the only possible (or no) palatalization for other consonants. I use
this marker also for palatalization {\em c} $\rightarrow$ {\em è} in
vocative sg of the paradigm {\em chlapec}. The final {\em c} is in
fact a palatalized {\em k}, so there is even a linguistic motivation
for this.
\item[\^{}2P] --- 2nd palatalization for {\em g, h} and {\em k}, or
the only possible (or no) palatalization for other consonants.
\item[\^{}A] --- Assimilation (or nothing).
\item[\^{}N] --- No alternation.
\end{description}

These markers are followed by a number that denotes how many
characters of the base form should be removed before attaching the
ending/suffix. Thus there are markers \^{}1P0, \^{}2P0, \^{}1P1,
etc. The markers starting with \^{}N only denote the length of the
ending of the base form---and instead of using \^{}N0 I attach the
suffix/ending directly to the base form.  Fortunately, nearly all
paradigms and derivations cause at most one type of alternation, so it
is possible to use one marker for the whole paradigm.

The markers for an epenthetic {\em e} are \^{}E1 (for {\em e} that
should be deleted) and \^{}E2 (for {\em e} that should be
inserted). The marker for the suffix {\em -ík} in derivations is
\^{}IK.

Here are some examples of lexical items and the rules that transduce
them to the surface form: \\

\noindent
(1) {\tt doktorka\^{}1P1in\^{}2P0ých} \\ The base form is {\em
doktorka} `she-doctor'. The marker \^{}1P1 denotes that the possible
alternation at this morpheme boundary is (first) palatalization and
that the length of the ending of the base form is 1 (it means that
{\em a} must be removed from the word form and the possible
alternation concerns {\em k}). The marker \^{}2P0 means that the
derived possessive adjective has a \mbox{\O-ending} and the possible
alternation at this morpheme boundary is palatalization. If we rewrite
this string to a sequence of morphemes we get the following string:
{\em doktork-in-ých}. The sound {\em k} in front of {\em i} is
palatalized, so the correct final form is {\em doktorèiných}, which is
genitive plural of the possessive adjective derived from the word {\em
doktorka}.

Let us look now at the two-level rules that transduce the lexical
string to the surface string. We need four rules in this example: two
for deleting the markers, one for deleting the ending {\em -a}, and
one for palatalization. The rules for deleting auxiliary markers are
very simple, as these markers should be deleted in any context. The
rules can be included in the definition of the alphabet of symbols:

\small
\begin{verbatim}
Alphabet
%^1P0:0 %^1P1:0
%^2P0:0 %^2P1:0 %^2P2:0 %^2P3:0
%^A2:0
%^N1:0 %^N2:0 %^N3:0 %^N4:0
%^E1:0 %^E2:0 %^IK:0
\end{verbatim} \normalsize
This notation means that the auxiliary markers are always realized
as zeros on the surface level. 

The rule for deleting the ending {\em -a} looks as follows:

\small
\begin{verbatim}
"Deletion of the ending -a-"
a:0 <=> _ [ %^N1: | %^1P1: | %^2P1: ] ;
        _ t: [ %^N2: | %^N4: ];
\end{verbatim} \normalsize

The first line of the rule describes the context of a one-letter
nominal ending {\em a}, and the second line describes the context of an
infinitive suffix with ending {\em -at} or {\em -ovat}.

The rule for palatalization {\em k} $\rightarrow$ {\em è} looks as
follows:

\small
\begin{verbatim}
"First palatalization k -> è"		
k:è <=> _ (%^IK:) [ a: | ì: ] %^1P1: i ;
        NonCÈS: _ (End) %^1P1: ì: ;
\end{verbatim} \normalsize

The first line describes two possible cases: either the derivation of
a possesive adjective from a feminine noun ({\em doktorka}
$\rightarrow$ {\em doktorèin}), or the derivation of a possesive
adjective from a feminine noun derived from a masculine that ends with
{\em -ík} ({\em úøedník} $\rightarrow$ ({\em úøednice} $\rightarrow$)
{\em úøednièin}).

The second context describes a comparative of an adjective, or a
comparative of adverb derived from that adjective ({\em hoøký}
$\rightarrow$ {\em hoøèej¹í}, {\em hoøèeji}). The set {\tt NonCÈS}
contains all character except {\em c, è} and {\em s} and it is
defined in a special section. This context condition is introduced,
because the groups of consonants {\em ck, èk} and {\em sk} have
different 1st palatalization.

The label {\tt End} denotes any character that can occur in an ending
and that is removed from the base form. \\

\noindent
(2) {\tt korek\^{}2P0\^{}E1em} \\ The base form of this word
form is {\em korek} `cork'; the marker \^{}2P0 means that the possible
alternation is (second) palatalization and that the length of ending
of the base form is 0. The marker \^{}E1 means that the base form
contains an epenthetic {\em e}, and {\em em} is the ending of
instrumental singular. The correct final form is {\em korkem}. The
rule for deleting an (epenthetic) {\em e} follows:

\small
\begin{verbatim}
"Deletion of e"
e:0 <=> Cons _ c: %^N2:;
    _ [ %^1P1: | %^2P1: | %^N1: | %^N2: ];
    Cons _ Cons: ([%^1P0:|%^2P0:]) %^E1: Vowel:;
    _ t:0 [ %^2P2: | %^N2: ];
\end{verbatim} \normalsize

The first line describes the context for deletion of the suffix {\em
-ec} in the derivation of the type {\em vìdec} `scientist'
$\rightarrow$ {\em vìdkynì} `she-scientist'.

The second context is the context of the ending {\em -e} or the suffix
{\em -ce}. This suffix must be removed in the derivation of the type
{\em soudce} `judge' $\rightarrow$ {\em soudkynì} `she-judge'.

The third context is the context of an epenthetic {\em e} that is
present in the base form and must be removed from a form with a
\mbox{non-\O{}} ending. The sets {\tt Cons} and {\tt Vowel} contain
all consonants and all vowels, respectively.

The fourth line describes the context for deletion of the infinitive
ending {\em -et}. \\

The whole program contains 35 rules. Some of the rules concern rather
morphology than phonology; namely the rules that remove endings or
suffixes. One rule is purely technical; it is one of the two rules
for the alternation {\em ch}~$\rightarrow$~{\em ¹}, as {\em c} and
{\em h} must be treated separately (though {\em ch} is considered one
letter in Czech alphabet). Six rules are forced by the Czech spelling
rules (e.g. rules for treating /{\em ï}/, /{\em »}/ and /{\em ò}/ in
various contexts, or a rule for rewriting {\em y} $\rightarrow$ {\em
i} after soft consonants). 18 rules deal with the actual phonological
alternations and they cover the whole productive phonological system
of Czech language. The lexicon using these rules was tested on a
newspaper text containing 2,978,320 word forms, with the result of
more than 96\% analyzed forms.

\section{Acknowledgements}

My thanks to Ken Beesley, who taught me how to work with the Xerox
tools, and to my father, Jan Skoumal, for fruitful discussions on the
draft of this paper.


\end{document}